\documentclass{article}

\usepackage[preprint]{neurips_2026}

\usepackage[utf8]{inputenc}
\usepackage[T1]{fontenc}
\usepackage{hyperref}
\usepackage{url}
\usepackage{booktabs}
\usepackage{amsfonts}
\usepackage{amsmath}
\usepackage{amssymb}
\usepackage{amsthm}
\usepackage{bbm}
\usepackage{nicefrac}
\usepackage{microtype}
\usepackage{xcolor}
\usepackage{multirow}
\usepackage{colortbl}
\usepackage{graphicx}
\usepackage{caption}
\usepackage{subcaption}
\usepackage{enumitem}
\usepackage{wrapfig}
\usepackage{makecell}
\usepackage{float}
\usepackage{kotex}

\newtheorem{proposition}{Proposition}
\newcommand{\sdev}[1]{{\scriptsize $\pm$\,#1}}

\newcommand{\nj}[1]{\textcolor{black}{#1}}

\title{Advancing Entropy-Level Credit Assignment in RLVR via Proximal Entropy Policy Optimization\thanks{Accepted at NeurIPS 2026.}}

\author{%
  Yun Kim \\
  Seoul National University \\
  \texttt{yunkimmy@snu.ac.kr}
  \And
  Nojun Kwak \\
  Seoul National University \\
  \texttt{nojunk@snu.ac.kr}
}

\begin{document}

\maketitle

\begin{abstract}
    Value-model-free RLVR methods such as GRPO assign uniform advantages to all tokens in a rollout, ignoring that tokens contribute unequally. Recent methods use token entropy as an importance proxy but compute it globally across the batch, conflating importance with prompt difficulty and positional trends. We argue that importance should instead be measured relative to a token's own local context, and introduce proximal entropy: a local measure of token importance relative to neighboring tokens, and prove it is invariant to both confounders.  Proximal Entropy Policy Optimization (PEPO) uses it to weight per-token advantages and outperforms GRPO and entropy-based baselines on mathematical reasoning across Qwen3-1.7B, Qwen3-4B, and Llama-3.2-3B-Instruct. We also show the formulation generalizes to other algorithms where substituting proximal entropy into existing methods improves, and applying it to single-stream RL succeeds where global entropy fails.
\end{abstract}

\section{Introduction}

In Reinforcement Learning with Verifiable Rewards (RLVR)~\cite{lambert2024tulu, guo2025deepseekr1} for large language models, value-model-free methods such as Group Relative Policy Optimization (GRPO)~\cite{shao2024deepseekmath} have become dominant for their efficiency. However, these methods assign a single sequence-level advantage to every generated token, ignoring the fact that tokens contribute unequally to the final answer. Several approaches address this through learned value models~\cite{schulman2017ppo, grpolambda2025} or additional rollouts~\cite{kazemnejad2025vineppo, wang2025segmentspo}, but these require training a separate model or generating extra samples, adding significant overhead. Among lighter alternatives, token entropy has emerged as the most widely adopted proxy for token importance~\cite{wang2025beyond, cheng2025reasoning, tan2025gtpo}, requiring no overhead beyond the standard forward pass.

The premise of entropy-based credit assignment is that low-entropy tokens correspond to routine grammatical continuations, while high-entropy tokens mark \emph{pivotal decisions} in the reasoning trajectory~\cite{wang2025beyond}. Methods such as 80/20~\cite{wang2025beyond} exploit this by updating only the top 20\% of tokens by entropy, ranked across all tokens in a training batch. However, we observe that this batch-level approach systematically biases updates toward hard prompts and early generation steps, where entropy is globally elevated. This causes low-importance tokens in hard prompts or early positions to be prioritized over genuine decision points, misguiding the learning signal. Therefore, we argue that a token's importance should be measured relative to its own \emph{local context}, not relative to the batch.

In this paper, we propose Proximal Entropy Policy Optimization (PEPO), which assigns each token an advantage weight based on its \textit{proximal entropy}. Proximal entropy measures how uncertain the model is at a given position relative to its neighboring tokens, rather than relative to the entire batch. We define proximal entropy as a softmax over a sliding window of entropies centered on each token, providing a per-token importance signal that is invariant to prompt difficulty and robust to positional trends. We formally show that this formulation removes both confounders (Proposition~\ref{prop:invariance}) and empirically verify that proximal entropy remains constant across difficulty levels and generation positions where global entropy varies substantially. Importantly, computing proximal entropy adds negligible overhead, as token-level entropy is precomputed by standard LLM inference engines and the sliding window operations are fully vectorized.

\begin{figure}[t!]
    \centering
    \begin{subfigure}{0.49\textwidth}
        \centering
        \includegraphics[width=\linewidth]{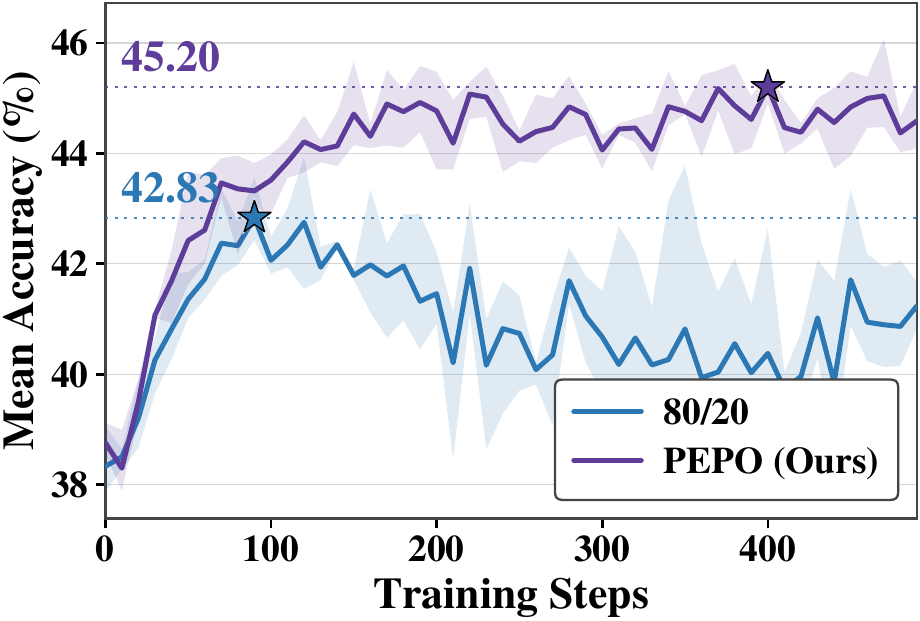}
        \caption{Qwen3-1.7B}
        \label{fig:qwen3_1.7b}
    \end{subfigure}
    \hfill
    \begin{subfigure}{0.49\textwidth}
        \centering
        \includegraphics[width=\linewidth]{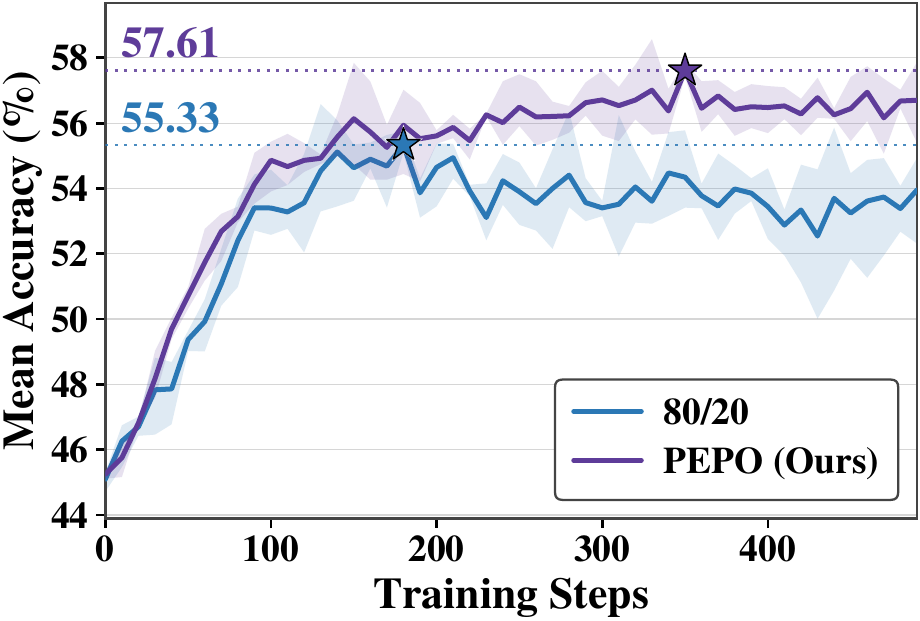}
        \caption{Qwen3-4B}
        \label{fig:qwen3_4b}
    \end{subfigure}
    \caption{Validation accuracy over training steps for PEPO and the 80/20. The plotted values represent the average mean@k accuracy evaluated across the MATH, AIME 2024, AIME 2025, and AMC benchmarks for Qwen3-1.7B (a) and Qwen3-4B (b). PEPO achieves a higher validation accuracy compared to the 80/20 baseline.}
    \label{fig:training_graphs}
\end{figure}

We conduct extensive experiments on MATH500~\cite{lightman2023letsverify}, AMC~\cite{maa_amc23}, and AIME 2024/2025~\cite{maa_aime2024} using Qwen3-1.7B, Qwen3-4B~\cite{yang2025qwen3}, and Llama-3.2-3B-Instruct~\cite{grattafiori2024llama}, where PEPO consistently outperforms batch-level methods on Qwen3-1.7B and Qwen3-4B (Figure~\ref{fig:training_graphs}). Beyond accuracy gains, we find that the proximal entropy formulation is itself the key ingredient. Substituting our local reference frame into 80/20 improves its performance without any other changes, isolating the contribution from the rest of PEPO's design. The same formulation also transfers to single-stream RL (SPO)~\cite{xu2025spo}, where each prompt produces only a single rollout. In this setting, the diversity of prompts greatly increases, and batch-level entropy methods degrade below the baseline while PEPO continues to improve over it. Together, these results confirm that proximal entropy captures token importance more effectively than batch-level approaches, providing accurate credit assignment for value-model-free RL. We summarize our contributions as follows.

\begin{itemize}
    \item We identify a systematic bias in batch-level entropy metrics~\cite{wang2025beyond, cheng2025reasoning} that concentrates updates on hard prompts and early generation steps.
    \item We propose PEPO, which weights per-token advantages using \textit{proximal entropy}, a local measure of token importance that we formally prove is invariant to prompt difficulty and  positional trends.
    \item We demonstrate that PEPO consistently outperforms batch-level methods across three model families \nj{(Qwen3-1.7B, Qwen3-4B, and Llama-3.2-3B-Instruct)} on \nj{four benchmarks (MATH500, AMC, AIME 2024, and AIME 2025).}
    \item We show that the proximal entropy formulation generalizes beyond PEPO, improving 80/20 when substituted in and transferring to single-stream RL where batch-level entropy fails.
\end{itemize}

\section{Related Work}
\label{sec:related}

\paragraph{Value-model-free RLVR and credit assignment.}
GRPO~\citep{shao2024deepseekmath} and its successors~\citep{yu2025dapo, zheng2025gspo, zhao2025gmpo, chen2025cispo} eliminate the value network of PPO by estimating advantages through group-relative comparisons, and Single-stream Policy Optimization~\citep{xu2025spo} further removes the group synchronization barrier.
These methods differ in how they estimate and stabilize advantages but share a common limitation that every token in a rollout receives identical credit.
A range of approaches address this through auxiliary signals.
Learned value networks~\citep{schulman2017ppo, grpolambda2025}, Monte Carlo estimates from extra rollouts or tree-structured branches~\citep{kazemnejad2025vineppo, wang2025segmentspo, tran2025tempo, rtmc2026, ssvpo2025}, process reward models that score each step~\citep{pure2025, capo2025, reasonfluxprm2025}, attention maps from selected heads~\citep{li2025preplananchor, atrl2026, attnpo2026}, and gradient-based attribution~\citep{grad2reward2026} all recover stronger credit signals at the cost of additional models, extra rollouts, or extra computation per token. Token entropy, by contrast, is computed from the same forward pass as the policy itself and adds essentially zero overhead, making it the lightweight signal we focus on in this work.

\paragraph{Token-entropy-based credit assignment.}
Token entropy has emerged as the dominant lightweight proxy for token importance, since it requires no overhead beyond the standard forward pass.
\citet{wang2025beyond} show that high-entropy tokens correspond to forking points in reasoning trajectories and propose 80/20, which restricts updates to the top 20\% of tokens by entropy.
\citet{cheng2025reasoning} augment the advantage with a clipped, gradient-detached entropy term to amplify updates at high-entropy positions.
GTPO~\citep{tan2025gtpo} redistributes rewards by weighting each token by its entropy relative to other rollouts in its group, while UCAS~\citep{xie2025ucas} modulates advantages using both response-level confidence and token-level certainty.
ERPO~\citep{yu2026erpo} and EDGE-GRPO~\citep{zhang2025edgegrpo} further explore entropy-aware gating and advantage diversity.
A common thread across these methods is that entropy statistics are computed in a global reference frame, either across the entire batch or as absolute per-token quantities.
As we demonstrate in Section~\ref{sec:entropy-signal}, this global frame conflates token importance with prompt difficulty and positional trends, leading to systematic biases in which tokens receive credit.

Closest to our work, ARES~\citep{chen2026ares} uses window-averaged entropy to shape per-token advantages in multimodal reasoning, adding a bonus on tokens whose window entropy exceeds a batch-derived threshold.
PEPO differs in three respects.
First, ARES uses a forward-looking window with arithmetic averaging to \emph{smooth} entropy and thresholds it against a batch-global cutoff, whereas PEPO uses a centered window with softmax-relative weighting that measures each token \emph{against} its local neighbors and is therefore invariant to the absolute scale.
Second, the symmetric window is essential to our positional invariance result (Proposition~\ref{prop:invariance}), since the cancellation of linear positional trends in the proof relies on offsets ranging symmetrically around $t$; a forward-looking window would retain a residual trend bias.
Third, we show that the local frame is a drop-in improvement that transfers across algorithms and to the single-stream setting where global entropy methods fail.
\section{Background and Motivation}
\label{sec:background}

\subsection{Preliminaries}
\label{sec:prelim}
We consider the RLVR setting~\citep{lambert2024tulu, guo2025deepseekr1}, in which a policy $\pi_\theta$ generates a response $y = (y_1, \dots, y_T)$ conditioned on a prompt $x$, and a verifier assigns a binary reward $r(x, y) \in \{0, 1\}$ indicating whether $y$ is correct. The goal is to maximize the expected reward $\mathbb{E}_{x \sim p_{data}, y \sim \pi_\theta (\cdot|x)}[r(x,y)]$.
\paragraph{GRPO.} For each prompt $x$, GRPO~\citep{shao2024deepseekmath} samples a group of $G$ rollouts $\{y_i\}_{i=1}^G$ from the current policy and computes a group-normalized advantage
\begin{equation}
    A_i = \frac{r_i - \text{mean}(\{r_j\}_{j=1}^G)}{\text{std}(\{r_j\}_{j=1}^G)},
\end{equation}
where $r_i = r(x, y_i)$. The same advantage $A_i$ is assigned to every token $y_{i,t}$ in the rollout, and the policy is updated by maximizing
\begin{equation}
    \mathcal{J}_{\text{group}}(\theta) = \frac{1}{\sum_{j=1}^{G} T_j} \sum_{i=1}^{G} \sum_{t=1}^{T_i}
    \min\!\big( \rho_{i,t}(\theta) A_i,\, \text{clip}(\rho_{i,t}(\theta), 1-\varepsilon_l, 1+\varepsilon_h) A_i \big),
    \label{eq:GRPO}
\end{equation}
where \nj{$T_i$ is the number of tokens generated in the $i$-th rollout,} $\rho_{i,t}(\theta) = \pi_\theta(y_{i,t} \mid x, y_{i,<t}) / \pi_{\theta_{\text{old}}}(y_{i,t} \mid x, y_{i,<t})$ is the importance ratio and $\varepsilon_l, \varepsilon_h$ are the lower and upper clipping thresholds~\citep{schulman2017ppo, yu2025dapo}.
\paragraph{Token entropy.} The entropy of the policy at step $t$ of rollout $i$ is
\begin{equation}
    H_{i,t} = -\sum_{v \in \mathcal{V}} \pi_\theta(v \mid x, y_{i,<t}) \log \pi_\theta(v \mid x, y_{i,<t}),
\end{equation}
where $\mathcal{V}$ is the vocabulary.

\paragraph{Fine-grained credit assignment.} Because $A_i$ in Eq.~(1) is constant across $t$, GRPO assigns identical credit to every token in a rollout, regardless of its role in producing $r_i$. We refer to the problem of assigning per-token weights $c_{i,t}$, such that the effective advantage $c_{i,t} A_i$ reflects each token's contribution, as \emph{fine-grained credit assignment}. The challenge is that $r_i$ provides no direct signal about which tokens mattered, so any solution must rely on a proxy for token importance. A learned value model can provide such a proxy~\citep{schulman2017ppo, grpolambda2025}, but introduces substantial memory and training overhead and reintroduces the instability issues that motivated value-free RLVR in the first place~\citep{shao2024deepseekmath, yu2025dapo}. In this work, we explore token entropy as the importance signal, since it is a lightweight proxy that introduces no additional overhead, as it is already computed during standard LLM inference~\citep{kwon2023vllm}.

\subsection{Entropy as a Token Importance Signal}
\label{sec:entropy-signal}

\begin{wrapfigure}{r}{0.47\textwidth}
  \vspace{-5mm}
  \centering
  \includegraphics[width=\linewidth]{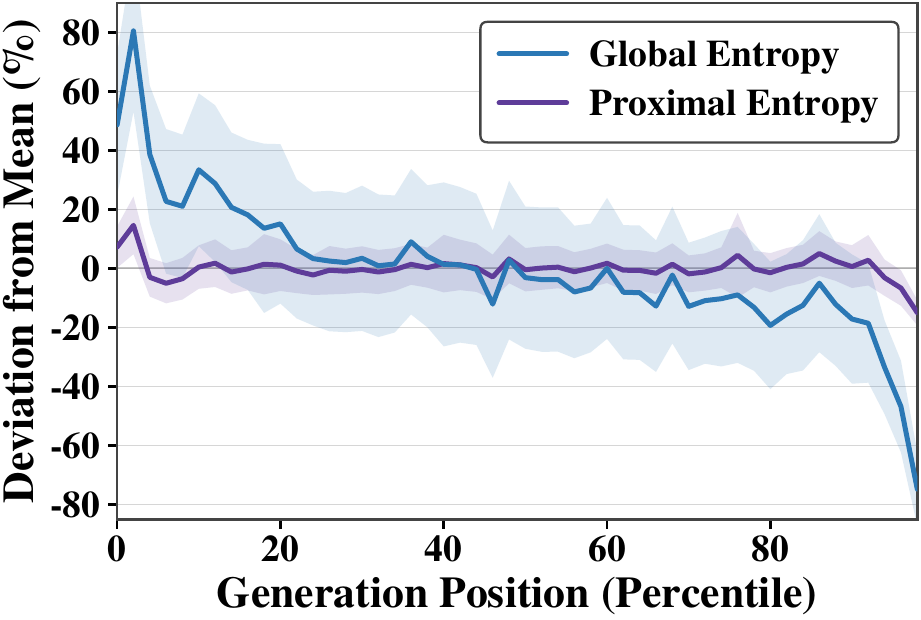}
  \caption{Positional bias in token entropy on Llama-3.2-3B-Instruct. We plot the deviation of global and proximal entropy from its sequence-level mean across generation percentiles.}
  \label{fig:entropy_position}
\end{wrapfigure}



Token entropy has emerged as a model-free alternative to learned value models~\citep{wang2025beyond, cheng2025reasoning, tan2025gtpo}. During generation, routine continuations such as grammatical placeholders and function words carry little uncertainty, while tokens at logical branch points, intermediate numerical results, and operator choices exhibit high entropy and frequently determine whether a derivation proceeds correctly~\citep{wang2025beyond}. A growing line of work exploits this by using entropy to instantiate the per-token weight $c_{i,t}$~\citep{wang2025beyond, cheng2025reasoning, xie2025ucas}. For instance, \citet{wang2025beyond} show that high-entropy tokens correspond to \emph{forking} points in the reasoning trajectory and propose 80/20, which updates only the top $20\%$ of tokens by entropy:
\begin{equation}
    c_{i,t} = \mathbbm{1}\left[ H_{i,t} \geq \tau_{80} \right], 
\end{equation}
where $\tau_{80}$ is the 80th-percentile entropy across all tokens in the batch. \citet{cheng2025reasoning} similarly augments the advantage function with a global entropy term to reinforce exploratory reasoning behaviors. In both cases, the reference frame is global.

\begin{wraptable}{r}{4.5cm}
\centering
\small
\vspace{-\intextsep}
\caption{Prompt difficulty bias in token entropy on Llama-3.2-3B-Instruct.}
\label{tab:bias}
\setlength{\tabcolsep}{4pt}
\begin{tabular}{lcc}
\toprule
 & \begin{tabular}{@{}c@{}}Global \\ Entropy\end{tabular} & \begin{tabular}{@{}c@{}}Proximal \\ Entropy\end{tabular} \\
\midrule
Hard & 0.315 & 0.00987 \\
Moderate & 0.270 & 0.00987 \\
Easy & 0.221 & 0.00989 \\
\bottomrule
\end{tabular}
\end{wraptable}

However, a token's raw entropy is driven not only by its importance but also by two confounding factors: the token's position in the sequence and the difficulty of the prompt. Figure~\ref{fig:entropy_position} illustrates the positional bias on Llama-3.2-3B-Instruct by plotting the deviation of global entropy from its sequence-level mean across generation percentiles, computed over 1024 rollouts from the validation set. Global entropy exhibits a strong positional trend, with early tokens deviating up to $80\%$ above the mean and late tokens falling well below it, reflecting that early generation steps carry higher uncertainty before the model commits to a reasoning direction~\citep{guo2025entropyreduction}. Table~\ref{tab:bias} illustrates the prompt difficulty bias. We group prompts by pass rate (out of $8$) into hard (0--1), moderate (2--6), and easy (7--8) categories and report the mean entropy for each group. Global entropy varies substantially across difficulty levels (0.315 for hard vs.\ 0.221 for easy), confirming that prompt difficulty elevates entropy across all tokens in a sequence. Together, these results demonstrate that batch-level entropy statistics systematically prioritize tokens from hard prompts and early positions, regardless of whether those tokens represent genuine decision points.

To understand why, consider the effect of prompt difficulty alone. Let $\bar{H}_i = \frac{1}{T_i}\sum_{t=1}^{T_i} H_{i,t}$ denote the mean entropy of rollout $i$, which serves as a measure of overall prompt difficulty. Each token's entropy decomposes as
\begin{equation}
    H_{i,t} = \underbrace{\bar{H}_i}_{\text{sequence-level}} + \underbrace{(H_{i,t} - \bar{H}_i)}_{\text{token-specific}},
\end{equation}
where the token-specific component measures how surprising a token is relative to its own sequence, serving as a more reliable indicator of importance. Global methods, however, rank tokens by $H_{i,t}$ across the entire batch, allowing tokens to receive disproportionate credit simply because they belong to a high-entropy sequence. Accounting for the positional bias only compounds this problem, as the systematic decay of entropy over the course of generation means that early tokens receive additional unwarranted credit due to their position alone.

These observations motivate a token importance measure that accounts for both prompt difficulty and positional variation. Rather than asking \emph{which tokens have high entropy relative to the global distribution}, we should ask \emph{which tokens have high entropy relative to their own local context}. In the next section, we formalize this local reference frame and introduce PEPO, which replaces global entropy statistics with a sliding window centered around each token.

\section{Proximal Entropy Policy Optimization}
\label{sec:method}

The key idea behind PEPO is to replace the global entropy statistics used by prior methods with a local measure we call \emph{proximal entropy}. For each token, proximal entropy captures how uncertain the model is at that position relative to its neighboring tokens, rather than relative to the entire batch. This provides a per-token importance signal that is invariant to prompt difficulty and position.

\paragraph{Proximal entropy.} Given the entropy sequence $(H_{i,1}, \dots, H_{i,T_i})$ for rollout $i$, we define the proximal entropy of token $t$ using a softmax over a sliding window of size $W$ centered on $t$:
\begin{equation}
    e_{i,t} = \frac{\exp(H_{i,t})}{\sum_{k \in \mathcal{W}(t)} \exp(H_{i,k})},
    \label{eq:e}
\end{equation}
where $\mathcal{W}(t) = \{k : |k - t| \leq W/2\}$ is the set of token positions within the window. The softmax formulation ensures that a token's proximal entropy reflects its uncertainty relative to its neighbors, not its absolute value. A token with moderately high entropy in an otherwise low-entropy region receives a large proximal entropy, while a token with the same entropy surrounded by equally uncertain neighbors does not.

\paragraph{Per-token advantage weighting.} We use proximal entropy to modulate the GRPO advantage. Since each token's proximal entropy is computed over its own local window, the values do not sum to one across the sequence. We therefore normalize and scale them to preserve the total advantage magnitude:
\begin{equation}
    \hat{A}_{i,t} = c_{i,t} \cdot A_i, \quad \text{where} \quad c_{i,t} = \frac{e_{i,t}}{\sum_{k=1}^{T_i} e_{i,k}} \cdot T_i.
    \label{eq:advantage}
\end{equation}
This ensures that $\sum_t c_{i,t} = T_i$, so the total advantage magnitude is identical to \nj{that of} standard GRPO. In practice, we apply proximal-entropy weighting only to positive-reward rollouts, consistent with prior entropy-weighted rewards for successful responses~\citep{tan2025gtpo} and asymmetric weighting of positive and negative samples~\citep{ma2026fipo}. We leave failed rollouts unweighted because amplifying their negative updates at high-entropy positions may discourage exploration. The PEPO objective is then
\begin{equation}
    \mathcal{J}_{\text{PEPO}}(\theta) = \frac{1}{\sum_{j=1}^{G} T_j} \sum_{i=1}^{G} \sum_{t=1}^{T_i}
    \min\!\big( \rho_{i,t}(\theta)\, \hat{A}_{i,t},\, \text{clip}(\rho_{i,t}(\theta), 1-\varepsilon_l, 1+\varepsilon_h)\, \hat{A}_{i,t} \big).
\end{equation}
This is identical to the standard GRPO objective in Eq.~\nj{(\ref{eq:GRPO})}, with the only difference being that the uniform advantage $A_i$ is replaced by the entropy-weighted $\hat{A}_{i,t}$.

\paragraph{Theoretical justification.} We now show that proximal entropy formally removes the confounders identified in Section~\ref{sec:entropy-signal}.
\begin{proposition}
\label{prop:invariance}
The proximal entropy $e_{i,t}$ \nj{in Eq.~(\ref{eq:e})} satisfies:
\begin{enumerate}[label=(\roman*)]
    \item If all entropies in a sequence are shifted by a sequence-dependent constant $\delta_i$, then $e_{i,t}$ is unchanged.
    \item If entropies in a sequence follow a smooth monotonic trend, then $e_{i,t}$ is approximately invariant to the trend for sufficiently small $W$.
\end{enumerate}
\end{proposition}

Property (i) implies that proximal entropy is invariant to prompt difficulty, as all tokens in a sequence share the same sequence-level component $\bar{H}_i$. Property (ii) implies robustness to positional entropy trends, which manifest as a smooth decay over the course of generation. Together, these properties ensure that proximal entropy isolates the token-specific signal from both confounders. The full proof is provided in Appendix~\ref{app:proof}. 

We note that Figure~\ref{fig:entropy_position} and Table~\ref{tab:bias} empirically corroborate both properties. As shown in Figure~\ref{fig:entropy_position}, global entropy deviates by up to 80 points from its sequence-level mean across generation percentiles, while proximal entropy remains flat throughout, confirming the positional invariance of property (ii). Table~\ref{tab:bias} further shows that global entropy varies from 0.221 (easy) to 0.315 (hard) across prompt difficulty groups, whereas proximal entropy remains constant at $\approx$0.00988 across all three groups, confirming the shift invariance of property (i).
\section{Experiments}
\label{sec:experiments}
\begin{table}[t]
\centering
\caption{Main results on mathematical reasoning benchmarks. Trained methods are reported as mean $\pm$ sample standard deviation (SD) over three independent runs. We report avg@1 on MATH500 and avg@16 on AMC, AIME 2024, and AIME 2025; the Mean column summarizes each run's four benchmark scores. Best results per model are in \textbf{bold}.}
\label{tab:main}
\resizebox{\linewidth}{!}{%
\begin{tabular}{llccccc}
\toprule
Model & Method & AIME2024 & AIME2025 & AMC & MATH500 & Mean \\
\midrule
\multirow{5}{*}{Qwen3-1.7B} & Base & 14.02 \sdev{0.82} & 17.58 \sdev{0.74} & 45.21 \sdev{0.65} & 75.53 \sdev{0.48} & 38.09 \sdev{0.42} \\
& GRPO & 16.58 \sdev{0.44} & 18.65 \sdev{0.62} & 50.28 \sdev{0.99} & 80.27 \sdev{1.01} & 41.45 \sdev{0.50} \\
& Entropy Adv. & 21.44 \sdev{1.81} & 19.48 \sdev{1.72} & 52.52 \sdev{1.06} & 81.40 \sdev{1.44} & 43.71 \sdev{0.52} \\
& 80/20 & 20.97 \sdev{1.48} & 20.83 \sdev{0.75} & 52.13 \sdev{0.44} & 80.73 \sdev{1.50} & 43.67 \sdev{0.30} \\
& PEPO (Ours) & \textbf{21.74 \sdev{1.06}} & \textbf{22.43 \sdev{0.67}} & \textbf{55.26 \sdev{1.11}} & \textbf{82.67 \sdev{0.31}} & \textbf{45.53 \sdev{0.46}} \\
\midrule
\multirow{5}{*}{Qwen3-4B} & Base & 25.62 \sdev{0.95} & 20.14 \sdev{0.81} & 55.39 \sdev{0.72} & 79.80 \sdev{0.54} & 45.24 \sdev{0.51} \\
& GRPO & 32.75 \sdev{0.54} & 23.54 \sdev{0.76} & 59.96 \sdev{0.68} & 84.33 \sdev{1.22} & 50.15 \sdev{0.43} \\
& Entropy Adv. & 35.88 \sdev{1.54} & 26.21 \sdev{1.22} & 67.83 \sdev{0.57} & 89.73 \sdev{0.23} & 54.91 \sdev{0.23} \\
& 80/20 & 35.35 \sdev{2.47} & 30.62 \sdev{0.21} & 67.25 \sdev{1.53} & 89.73 \sdev{1.15} & 55.74 \sdev{0.86} \\
& PEPO (Ours) & \textbf{39.24 \sdev{1.27}} & \textbf{31.81 \sdev{0.13}} & \textbf{69.58 \sdev{0.59}} & \textbf{91.40 \sdev{1.83}} & \textbf{58.01 \sdev{0.14}} \\
\midrule
\multirow{5}{*}{Llama-3.2-3B-Instruct} & Base & 5.69 \sdev{0.45} & 0.62 \sdev{0.18} & 21.49 \sdev{0.52} & 45.93 \sdev{0.61} & 18.43 \sdev{0.28} \\
& GRPO & 7.22 \sdev{0.84} & 0.90 \sdev{0.52} & 21.43 \sdev{0.57} & 47.67 \sdev{1.33} & 19.31 \sdev{0.20} \\
& Entropy Adv. & 8.26 \sdev{0.84} & 0.56 \sdev{0.24} & 23.64 \sdev{0.74} & 47.60 \sdev{0.60} & 20.02 \sdev{0.44} \\
& 80/20 & 8.60 \sdev{0.61} & 0.49 \sdev{0.12} & 22.57 \sdev{0.91} & 47.67 \sdev{0.50} & 19.83 \sdev{0.03} \\
& PEPO (Ours) & \textbf{9.10 \sdev{0.52}} & \textbf{1.46 \sdev{0.55}} & \textbf{24.12 \sdev{0.36}} & \textbf{49.33 \sdev{0.12}} & \textbf{21.00 \sdev{0.15}} \\
\bottomrule
\end{tabular}%
}
\end{table}

\subsection{Experimental Setup}
\label{sec:setup}

\paragraph{Models.} We evaluate on three base models spanning two model families: Qwen3-1.7B, Qwen3-4B~\citep{yang2025qwen3}, and Llama-3.2-3B-Instruct~\citep{grattafiori2024llama}.

\paragraph{Training.} For the main experiments, we use the ROLL framework~\citep{wang2025roll} with the DeepMath-103K dataset~\citep{he2025deepmath}. For single-stream RL experiments (Section~\ref{sec:spo}), we use the VeRL framework~\citep{sheng2024hybridflow} with the DAPO-Math-17K dataset~\citep{yu2025dapo}. All methods are implemented on top of GRPO~\citep{shao2024deepseekmath} with $G = 8$ rollouts per prompt, a learning rate of $1 \times 10^{-6}$, clipping threshold $\varepsilon_h = 0.28$, and $500$ training steps, and follow the default library configurations. For PEPO, we set the window size $W = 101$ across all experiments. The average response length during training ranges from approximately $2000$ to $3000$ tokens depending on the model, so the window covers roughly 3--5\% of each sequence. Full hyperparameter details are provided in Appendix.

\paragraph{Baselines.} We compare PEPO against three baselines: (1) standard GRPO~\citep{shao2024deepseekmath} with uniform token weighting, (2) 80/20~\citep{wang2025beyond}, which updates only the top $20\%$ of tokens by global entropy, and (3) Entropy-augmented advantage~\citep{cheng2025reasoning}, which augments the advantage function with a global entropy term.

\paragraph{Evaluation.} We evaluate on MATH500~\citep{lightman2023letsverify}, AMC~\citep{maa_amc23}, AIME 2024~\citep{maa_aime2024}, and AIME 2025~\citep{maa_aime2025}. We report avg@1 accuracy on MATH500 and avg@16 accuracy on AMC, AIME 2024, and AIME 2025 over three runs. All detailed per-run results are provided in the Appendix.

\subsection{Main Results}
\label{sec:main-results}
Table~\ref{tab:main} presents the main results across all models and benchmarks. PEPO consistently outperforms both GRPO and batch-level entropy methods across all three models spanning two distinct model families, demonstrating that the improvements are robust and not an artifact of a particular architecture.

On Qwen3-4B, PEPO achieves the largest improvements, outperforming GRPO by $6.49$\nj{\%p} on AIME 2024 and $9.62$\nj{\%p} on AMC, and surpassing 80/20 by $2.27$\nj{\%p} in mean accuracy across all benchmarks. On Qwen3-1.7B, PEPO improves over GRPO by $5.16$\nj{\%p} on AIME 2024 and $4.98$\nj{\%p} on AMC, with a mean accuracy gain of $1.82$\nj{\%p} over Entropy Advantage. Notably, on Llama-3.2-3B-Instruct, 80/20 and Entropy Adv. degrade below GRPO on AIME 2025 ($0.41\%$p and $0.34\%$p), while PEPO improves over GRPO on all benchmarks. The consistent gains of PEPO, even in cases where global entropy methods hurt performance, suggest that proximal entropy provides a more reliable token importance signal across model families.

\begin{table}[t]
\centering
\caption{Effect of substituting proximal entropy into 80/20. Each cell reports the mean $\pm$ sample SD over three independent runs. All other components of 80/20 remain unchanged. PEPO results from Table~\ref{tab:main} are included for reference. Best results within each model are bolded.}
\label{tab:proximal-80-20}
\small
\setlength{\tabcolsep}{3pt}
\begin{tabular}{llccccc}
\toprule
Model & Method & AIME2024 & AIME2025 & AMC & MATH500 & Mean \\
\midrule
\multirow{3}{*}{Qwen3-1.7B} & 80/20 & 20.97 \sdev{1.48} & 20.83 \sdev{0.75} & 52.13 \sdev{0.44} & 80.73 \sdev{1.50} & 43.67 \sdev{0.30} \\
& 80/20 + Proximal Entropy & \textbf{22.01 \sdev{0.87}} & 20.69 \sdev{0.24} & 53.36 \sdev{0.78} & 82.27 \sdev{0.31} & \textbf{44.58 \sdev{0.43}} \\
& PEPO (Ours) & 21.74 \sdev{1.06} & \textbf{22.43 \sdev{0.67}} & \textbf{55.26 \sdev{1.11}} & \textbf{82.67 \sdev{0.31}} & 45.53 \sdev{0.46} \\
\midrule
\multirow{3}{*}{Qwen3-4B} & 80/20 & 35.35 \sdev{2.47} & 30.62 \sdev{0.21} & 67.25 \sdev{1.53} & 89.73 \sdev{1.15} & 55.74 \sdev{0.86} \\
& 80/20 + Proximal Entropy & 36.18 \sdev{2.61} & 31.60 \sdev{0.48} & 69.38 \sdev{1.41} & 90.33 \sdev{0.31} & 56.87 \sdev{0.56} \\
& PEPO (Ours) & \textbf{39.24 \sdev{1.27}} & \textbf{31.81 \sdev{0.13}} & \textbf{69.58 \sdev{0.59}} & \textbf{91.40 \sdev{1.83}} & \textbf{58.01 \sdev{0.14}} \\
\bottomrule
\end{tabular}
\end{table}

\subsection{Proximal Entropy as a General Formulation}
\label{sec:general-formulation}

To isolate the contribution of the proximal entropy formulation from the rest of PEPO's design, we substitute our window-relative entropy into 80/20, replacing its global entropy ranking with proximal entropy while keeping all other components unchanged. Concretely, instead of selecting the top $20\%$ of tokens by global entropy across the batch, we select the top $20\%$ by proximal entropy within the same batch. This tests whether proximal entropy is indeed a more accurate measure of token importance compared to global entropy. Table~\ref{tab:proximal-80-20} shows the results.

Replacing global entropy with proximal entropy improves 80/20 across benchmarks on both models, with no other changes to the algorithm. This confirms that proximal entropy is a more accurate measure of token importance regardless of the algorithm it is embedded in. PEPO, which further replaces binary token selection with continuous entropy-based weighting, achieves the best overall performance across both models.

\subsection{Transfer to Single-Stream RL}
\label{sec:spo}

To test whether the proximal entropy formulation transfers beyond GRPO, we apply it to Single-stream Policy Optimization (SPO)~\citep{xu2025spo}, where each prompt produces only a single rollout. We refer to this variant as PESPO \nj{(Proximal Entropy SPO)}. We use the same window size $W = 101$ and all other hyperparameters as in the main experiments, without any task-specific tuning. Table~\ref{tab:spo} shows the results.
\begin{wraptable}{r}{0.6\linewidth}
\centering
\caption{Results on single-stream RL (SPO) with Qwen3-4B. Cells show the mean followed by sample SD in smaller type, computed over three independent runs. All methods are implemented on top of SPO. PESPO uses the same hyperparameters as in the main experiments without modification.}
\label{tab:spo}
\resizebox{\linewidth}{!}{%
\begin{tabular}{lccccc}
\toprule
Method & AIME2024 & AIME2025 & AMC & MATH500 & Mean \\
\midrule
SPO & \mbox{35.70 \sdev{1.97}} & \mbox{31.39 \sdev{0.44}} & \mbox{70.81 \sdev{1.28}} & \mbox{90.60 \sdev{0.53}} & \mbox{57.13 \sdev{0.42}} \\
S-Entropy Adv. & \mbox{38.89 \sdev{1.26}} & \mbox{21.80 \sdev{0.67}} & \mbox{68.42 \sdev{0.43}} & \mbox{89.47 \sdev{0.31}} & \mbox{54.65 \sdev{0.58}} \\
S-80/20 & \mbox{35.28 \sdev{1.15}} & \mbox{27.64 \sdev{1.05}} & \mbox{70.06 \sdev{0.93}} & \textbf{\mbox{91.00 \sdev{0.40}}} & \mbox{55.99 \sdev{0.43}} \\
PESPO (Ours) & \textbf{\mbox{41.74 \sdev{0.52}}} & \textbf{\mbox{31.60 \sdev{0.79}}} & \textbf{\mbox{71.71 \sdev{1.06}}} & \mbox{90.67 \sdev{0.31}} & \textbf{\mbox{58.93 \sdev{0.48}}} \\
\bottomrule
\end{tabular}%
}
\end{wraptable}

Both global entropy methods degrade below the SPO baseline in mean accuracy, with S-Entropy Adv. dropping to $54.65\%$ and S-80/20 to $55.99\%$, compared to SPO's $57.13\%$ in mean accuracy. The degradation is particularly pronounced on AIME 2025, where S-Entropy Adv. falls to $21.80\%$, nearly $10$\nj{\%p} below SPO. This is consistent with the failure mode identified in Section~\ref{sec:entropy-signal}. Without multiple rollouts per prompt, the batch contains a more diverse set of prompts, amplifying the prompt difficulty bias in global entropy statistics. PESPO, in contrast, achieves the highest mean accuracy ($58.93\%$) and improves over SPO on AIME 2024 by $6.04$\nj{\%p}, demonstrating that proximal entropy transfers to single-stream settings without hyperparameter adjustment. This provides further evidence that proximal entropy effectively captures token importance even when prompt diversity is large, whereas global entropy does not.

\subsection{Analysis}
\label{sec:analysis}

\paragraph{Token intervention.}
\begin{wraptable}{r}{0.49\textwidth}
\centering
\footnotesize
\caption{Token-replacement intervention on AIME 2024 using Qwen3-4B. Results are avg@16 accuracy (\%); each budget is the fraction of token positions replaced.}
\label{tab:token-intervention}
\setlength{\tabcolsep}{4pt}
\begin{tabular}{lrr}
\toprule
Budget & Global entropy & Proximal entropy \\
\midrule
0\%  & 25.62 & 25.62 \\
5\%  & 27.34 & \textbf{31.62} \\
10\% & 29.91 & \textbf{33.33} \\
15\% & 30.76 & \textbf{33.33} \\
20\% & 32.47 & \textbf{33.33} \\
30\% & \textbf{33.33} & \textbf{33.33} \\
\bottomrule
\end{tabular}
\end{wraptable}

To test the causal effect of the token-importance metrics on accuracy, we use each metric to choose which generated tokens to replace, following prior token-replacement analyses~\citep{meng2026sparse,huang2026direction}. For AIME 2024 avg@16, we generate responses with Qwen3-4B Base and rank their token positions by global entropy or proximal entropy. At each replacement budget, we replace the highest-ranked Base tokens with samples from a fixed GRPO checkpoint, then continue generation autoregressively. We hold the Base model, replacement checkpoint, evaluation set, and budget fixed, so only the ranking metric changes. Table~\ref{tab:token-intervention} reports the resulting accuracy.

At a 5\% replacement budget, proximal entropy reaches 31.62\% accuracy versus 27.34\% for global entropy, a 4.28-point gap. At 10\%, it reaches the GRPO checkpoint's 33.33\% accuracy; global entropy reaches that level only at 30\%. These results provide evidence that proximal entropy is a more accurate measure of token importance than global entropy in this setting, as it allows the GRPO checkpoint's full accuracy to be recovered with a smaller replacement budget, further supporting the claims in the main section.

\paragraph{Window size.} We ablate the window size $W$ on Qwen3-4B, comparing $W \in \{51, 101, 151\}$. Results are shown in Table~\ref{tab:ablation}. Performance is relatively robust across all three window sizes, suggesting that PEPO is not sensitive to this hyperparameter. This is consistent with the local smoothness assumption in Proposition 1. As long as the window is small enough relative to the sequence length, the positional trend is approximately cancelled. Among the three, $W=101$ achieves the best results, which we use as the default across all experiments.

\begin{table}[H]
\centering
\small
\caption{Ablation studies on window size on Qwen3-4B. Each cell shows the mean followed by sample SD in smaller type over three independent runs.}
\label{tab:ablation}
\begin{tabular}{lcccc}
\toprule
$W$ & AIME2024 & AIME2025 & AMC & MATH500 \\
\midrule
$W=51$  & \mbox{36.25 \sdev{1.50}} & \mbox{30.62 \sdev{1.30}} & \mbox{68.39 \sdev{0.59}} & \mbox{90.67 \sdev{0.42}} \\
$W=101$ & \mbox{39.24 \sdev{1.27}} & \mbox{31.81 \sdev{0.13}} & \mbox{69.58 \sdev{0.59}} & \mbox{91.40 \sdev{1.83}} \\
$W=151$ & \mbox{36.81 \sdev{0.24}} & \mbox{28.27 \sdev{1.26}} & \mbox{69.35 \sdev{0.49}} & \mbox{92.27 \sdev{0.46}} \\
\bottomrule
\end{tabular}
\end{table}

\begin{figure}[t!]
    \centering
    \includegraphics[width=\linewidth]{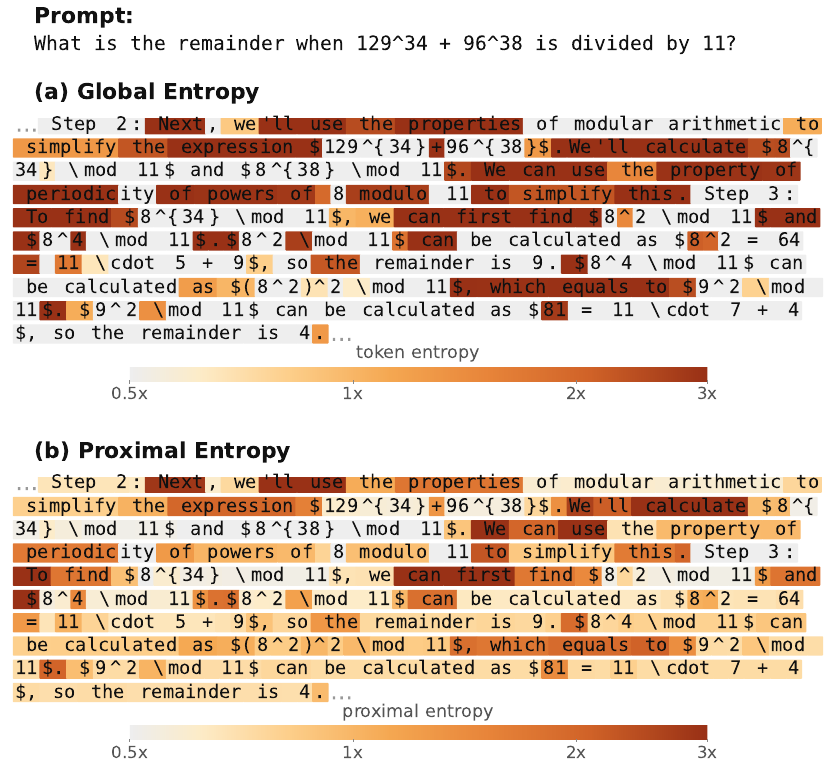}
    \caption{Qualitative comparison of token weighting on a Qwen3-4B rollout. \textbf{(a)}~Global entropy assigns broad, near-uniform weight across structural and decision tokens alike (e.g., the entire phrases ``We'll calculate'' and ``property of periodicity''). \textbf{(b)}~Proximal entropy evaluates each token relative to its local neighborhood, producing fine-grained sub-phrase weighting that singles out the tokens that steer the reasoning trajectory (e.g., \emph{calculate}, \emph{periodicity}) while still preserving signal across the derivation steps.}
    \label{fig:qualitative}
\end{figure}

\paragraph{Qualitative analysis.}
Figure~\ref{fig:qualitative} (Appendix) visualizes the token-level scores produced by global entropy and proximal entropy on a Qwen3-4B rollout for the modular arithmetic problem $129^{34} + 96^{38} \bmod 11$, where each token is colored by its score relative to the mean of each metric. Two properties of proximal entropy stand out. First, proximal entropy produces \emph{fine-grained, sub-phrase weighting} that global entropy cannot. In ``We'll calculate,'' for instance, global entropy lights up nearly uniformly across all three tokens, whereas proximal entropy distinguishes between them. \emph{We} and \emph{calculate} receive substantially more weight than \emph{'ll}, with \emph{calculate} receiving the strongest weight as the token that commits to the action. The same effect appears in ``property of periodicity,'' where global entropy assigns elevated weight to all three content words, while proximal entropy concentrates on \emph{periodicity}, the noun that names the mathematical principle being invoked. This local reference frame allows proximal entropy to identify the specific tokens that steer the reasoning trajectory, even within short phrases where every token shares similar local context.

Second, proximal entropy does not ignore the derivation itself. The execution tokens that carry out the modular arithmetic\,---\,``$8^2 = 64 = 11 \cdot 5 + 9$,'' ``$81 = 11 \cdot 7 + 4$''\,---\,receive visibly nontrivial weight under proximal entropy, even though their absolute entropy is near zero (each digit follows deterministically from the formula structure). Global entropy, in contrast, washes these tokens out entirely. Proximal entropy's window-relative formulation preserves signal across the derivation while still concentrating attention on the higher-leverage decision points.

Additional evaluation and ablations, including held-out code generation, the weighting design, and temperature sensitivity, are reported in Appendix~\ref{app:additional_experiments}.

\section{Conclusion}
\label{sec:conclusion}

We introduced Proximal Entropy Policy Optimization (PEPO), which replaces global entropy statistics with \textit{proximal entropy}, a local measure of token importance for value-model-free RLVR. By computing entropy relative to a sliding window around each token, PEPO identifies tokens that are genuinely important given their local context, rather than tokens that merely reside in high-entropy regions of the batch. We formally showed that this formulation is invariant to prompt difficulty and approximately invariant to positional entropy trends (Proposition~\ref{prop:invariance}), directly addressing the confounders we identified in global entropy methods. Experiments across Qwen3-1.7B, Qwen3-4B, and Llama-3.2-3B-Instruct demonstrated consistent improvements over GRPO and batch-level entropy methods on mathematical reasoning benchmarks. On a held-out LiveCodeBench split, Qwen3-4B with PEPO also improved pass@1 over GRPO and 80/20. We further showed that the proximal entropy formulation generalizes beyond PEPO itself. Substituting it into 80/20 yields gains without any other changes, and applying it to single-stream RL (PESPO) improves performance where global entropy degrades.

\paragraph{Limitations and future work.} Our main benchmark suite focuses on mathematical reasoning. The additional coding evaluation covers one held-out LiveCodeBench split and Qwen3-4B, so broader evaluation across coding benchmarks and model families is needed to establish how well PEPO generalizes to code generation, agentic tool use, and multi-turn dialogue. We did not compare against value-based methods such as PPO, which provide a stronger but more expensive form of credit assignment. Combining proximal entropy with a learned value model could offer complementary benefits. More broadly, the local reference frame could be applied beyond advantage weighting, for instance, to guide exploration strategies or to design entropy-aware curricula that adapt at the token level rather than the prompt level.

\begin{ack}
This work was supported by the Korean Government through grants from IITP (RS-2021-II211343 and RS-2025-25442338) and NRF (RS-2026-25500579).

\textbf{Competing interests:} The authors declare no competing interests.
\end{ack}

\bibliographystyle{plainnat}
\bibliography{custom}

\appendix

\clearpage
\appendix
\section*{Appendix}

\section{Broader Impacts}
\label{app:broader_impacts}

PEPO is a methodological contribution to value-model-free RLVR that improves
token-level credit assignment without additional rollouts, value models, or
process reward models. As a general-purpose post-training technique, its
societal impact is largely inherited from the broader trajectory of LLM
reasoning research rather than tied to a specific deployment.

On the positive side, PEPO reduces the compute required to obtain a given
level of reasoning performance: it matches or exceeds batch-level entropy
methods at the same training budget (Table~\ref{tab:runtime}) and transfers
to single-stream RL, where it removes the need for $G$ parallel rollouts per
prompt. Lower training cost lowers the barrier to entry for academic and
resource-constrained groups working on reasoning models, and reduces the
energy footprint of RL post-training.

The negative considerations are those common to any technique that improves
LLM reasoning. Stronger reasoning capabilities accelerate both beneficial
applications (scientific assistance, education, code generation) and
potentially harmful ones (more persuasive misinformation, automated
vulnerability discovery, misuse in high-stakes decisions). PEPO does not
introduce a new capability axis or unlock qualitatively new behaviors; it
makes existing RLVR pipelines more sample- and credit-efficient. We do not
release any new pretrained models or scraped datasets, and all experiments
build on publicly available base models and mathematical reasoning datasets
that have been released by their original authors with their own usage terms.
We see no direct path from this work to specific harmful applications beyond
those already discussed in the broader literature on LLM post-training.

\section{Proof of Proposition 1}
\label{app:proof}

We restate the proposition for convenience.

\begin{proposition}
The proximal entropy $e_{i,t}$ (Eq.~6) satisfies:
\begin{enumerate}[label=(\roman*)]
    \item If all entropies in a sequence are shifted by a sequence-dependent constant $\delta_i$, then $e_{i,t}$ is unchanged.
    \item If entropies in a sequence follow a smooth monotonic trend, then $e_{i,t}$ is approximately invariant to the trend for sufficiently small $W$.
\end{enumerate}
\end{proposition}

\subsection{Proof of (i): Invariance to Constant Shift}

Let $\tilde{H}_{i,t} = H_{i,t} + \delta_i$ for all $t$ and some sequence-dependent constant $\delta_i \in \mathbb{R}$.  The corresponding proximal entropy is
\begin{align}
    \tilde{e}_{i,t}
    &= \frac{\exp(\tilde{H}_{i,t})}{\sum_{k \in \mathcal{W}(t)} \exp(\tilde{H}_{i,k})}
    = \frac{\exp(H_{i,t} + \delta_i)}{\sum_{k \in \mathcal{W}(t)} \exp(H_{i,k} + \delta_i)} \notag \\
    &= \frac{\exp(\delta_i) \exp(H_{i,t})}{\exp(\delta_i) \sum_{k \in \mathcal{W}(t)} \exp(H_{i,k})}
    = \frac{\exp(H_{i,t})}{\sum_{k \in \mathcal{W}(t)} \exp(H_{i,k})}
    = e_{i,t}.
\end{align}
Since all tokens within a window $\mathcal{W}(t)$ belong to the same rollout $i$, they share the same $\delta_i$, which factors out and cancels.  This holds for any sequence-dependent shift, so proximal entropy is invariant to differences in overall entropy level across sequences.

\subsection{Proof of (ii): Approximate Invariance to Positional Trends}
We decompose each token's entropy as
\begin{equation}
    H_{i,t} = f(t) + r_{i,t},
\end{equation}
where $f \colon [T_i] \to \mathbb{R}$ is a smooth, monotonic, deterministic positional trend shared across rollouts and $r_{i,t}$ is the residual that captures token-specific variation. Our goal is to show that $e_{i,t}$ is approximately independent of $f$ under mild regularity conditions.

\paragraph{Setup.} Fix a token position $t$ and consider the window $\mathcal{W}(t) = \{k : |k - t| \leq W/2\}$. Since $f$ is monotonic, $f'(t)$ does not change sign within the window, so the linear term dominates the local variation of $f$ and a first-order Taylor expansion of $f$ around $t$ gives
\begin{equation}
    f(k) = f(t) + f'(t)(k - t) + \mathcal{O}\!\left(\frac{W^2}{T_i^2}\right),
    \label{eq:taylor}
\end{equation}
where we have used $|k - t| \leq W/2$ and the fact that $f$ varies over the full sequence length $T_i$, so $\sup|f''| = \mathcal{O}(1/T_i^2)$. Monotonicity ensures that the residual term captures only curvature rather than direction-changing oscillations, so the approximation is accurate when $W \ll T_i$, i.e., the positional trend is approximately linear within each window.

\paragraph{Substitution.} Substituting Eq.~\eqref{eq:taylor} into the definition of $e_{i,t}$ and dropping the higher-order terms:
\begin{align}
    e_{i,t}
    &= \frac{\exp(H_{i,t})}{\sum_{k \in \mathcal{W}(t)} \exp(H_{i,k})}
    = \frac{\exp\!\big(f(t) + r_{i,t}\big)}{\sum_{k \in \mathcal{W}(t)} \exp\!\big(f(k) + r_{i,k}\big)} \notag \\
    &\approx \frac{\exp\!\big(f(t) + r_{i,t}\big)}{\sum_{k \in \mathcal{W}(t)} \exp\!\big(f(t) + f'(t)(k-t) + r_{i,k}\big)} \notag \\
    &= \frac{\exp(r_{i,t})}{\sum_{k \in \mathcal{W}(t)} \exp\!\big(f'(t)(k-t) + r_{i,k}\big)},
    \label{eq:after_cancel}
\end{align}
where the last step uses the translation-invariance property established in (i) to cancel $f(t)$.

\paragraph{Small-gradient regime.} Eq.~\eqref{eq:after_cancel} still contains the linear term $f'(t)(k-t)$. We show that this term is negligible when the trend slope is small relative to the inverse window size. For each $k \in \mathcal{W}(t)$, we have $|f'(t)(k-t)| \leq |f'(t)| \cdot W/2$. When $|f'(t)| \cdot W/2 \ll 1$, a first-order expansion of the exponential gives
\begin{equation}
    \exp\!\big(f'(t)(k-t) + r_{i,k}\big) \approx \exp(r_{i,k})\big(1 + f'(t)(k-t)\big).
\end{equation}
Substituting into the denominator of Eq.~\eqref{eq:after_cancel}:
\begin{align}
    \sum_{k \in \mathcal{W}(t)} \exp\!\big(f'(t)(k-t) + r_{i,k}\big)
    &\approx \sum_{k \in \mathcal{W}(t)} \exp(r_{i,k}) + f'(t)\!\sum_{k \in \mathcal{W}(t)} (k-t)\exp(r_{i,k}).
    \label{eq:denom_expand}
\end{align}
The second term is a weighted sum of offsets $(k-t)$, which range symmetrically from $-W/2$ to $+W/2$. Even without any assumption on the residuals, this term is $\mathcal{O}(f'(t) \cdot W)$ times the first term, and therefore negligible when $|f'(t)| \cdot W/2 \ll 1$. The same expansion applies to the numerator at $k = t$, where $(k-t) = 0$ and the linear correction vanishes exactly. We therefore obtain
\begin{align}
    e_{i,t} &\approx \frac{\exp(r_{i,t})}{\sum_{k \in \mathcal{W}(t)} \exp(r_{i,k})},
    \label{eq:final}
\end{align}
which depends only on the residuals and is independent of the trend $f$.

\paragraph{Summary.} The conditions required are that the positional trend $f$ is smooth and monotonic, and that $|f'(t)| \cdot W/2 \ll 1$, i.e., the trend is slowly varying within each window. Under these conditions, the proximal entropy $e_{i,t}$ is approximately invariant to $f$ up to corrections of order $\mathcal{O}(f'(t) \cdot W)$ and the Taylor remainder from Eq.~\eqref{eq:taylor}. Combined with property (i), this implies that $e_{i,t}$ isolates the token-specific signal from both prompt difficulty and positional entropy trends. \qed

\paragraph{Remark.} The condition $|f'(t)| \cdot W/2 \ll 1$ is mild for typical reasoning rollouts. From Fig.~\ref{fig:entropy_position}, entropy decays from roughly 0.314 to 0.185 over the full sequence, giving $|f'| \approx 0.13/T_i$. For $T_i = 2000$ and $W = 101$, we have $|f'| \cdot W/2 \approx 0.004$, well within the small-gradient regime. The window size ablation in Fig~\ref{fig:window_bias} provides further empirical support, where performance is stable across $W \in \{31, 51, 101\}$, suggesting that the approximation holds well for the window sizes used.

\section{Implementation Details}
\label{app:impl}

\subsection{Hyperparameters}
\label{app:hyperparams}

We follow the default configurations of the ROLL framework~\citep{wang2025roll} for our main experiments and the VeRL framework~\citep{sheng2024hybridflow} for single-stream experiments. Table~\ref{tab:hyperparams_main} reports the hyperparameters used for the main and drop-in experiments (Tables 2, 3, 6), and Table~\ref{tab:hyperparams_spo} reports those used for the single-stream RL experiments (Table 4).

\begin{table}[H]
\centering
\caption{Hyperparameters for main experiments on ROLL framework. These settings are used for GRPO, 80/20, Entropy Adv., and PEPO across all three models.}
\label{tab:hyperparams_main}
\begin{tabular}{ll}
\toprule
\textbf{Parameter} & \textbf{Value} \\
\midrule
Framework & ROLL \\
Training data & DeepMath-103K \\
Rollouts per prompt ($G$) & 8 \\
Learning rate & $1 \times 10^{-6}$ \\
Optimizer & AdamW \\
Lower clipping threshold ($\varepsilon_l$) & 0.2 \\
Upper clipping threshold ($\varepsilon_h$) & 0.28 \\
Training steps & 500 \\
Sampling temperature & 0.6 \\
Top-$p$ & 0.6 \\
Maximum response length & 4096 tokens \\
Window size $W$ (PEPO) & 101 \\
\bottomrule
\end{tabular}
\end{table}

\begin{table}[H]
\centering
\caption{Hyperparameters for single-stream experiments on VeRL framework. These settings are used for SPO, S-80/20, S-Entropy Adv., and PESPO.}
\label{tab:hyperparams_spo}
\begin{tabular}{ll}
\toprule
\textbf{Parameter} & \textbf{Value} \\
\midrule
Framework & VeRL \\
Training data & DAPO-Math-17K \\
Rollouts per prompt & 1 (single-stream) \\
Learning rate & $1 \times 10^{-6}$ \\
Optimizer & AdamW \\
Lower clipping threshold ($\varepsilon_l$) & 0.2 \\
Upper clipping threshold ($\varepsilon_h$) & 0.28 \\
Training steps & 500 \\
Sampling temperature & 0.6 \\
Top-$p$ & 0.6 \\
Maximum response length & 4096 tokens \\
Window size $W$ (PESPO) & 101 \\
\bottomrule
\end{tabular}
\end{table}

\subsection{Window Edge Handling}
\label{app:reflect}

The window $\mathcal{W}(t) = \{k : |k - t| \leq W/2\}$ extends beyond the sequence boundaries when $t < W/2$ or $t > T_i - W/2$. We use reflect padding to fill the missing positions, mirroring entropy values across the boundary. We chose reflect padding over zero padding because the latter would inject artificially low entropy values into the softmax denominator at boundary windows, distorting the relative weighting. Reflect padding instead uses values drawn from the same sequence, preserving the local entropy scale. This is also intuitively aligned with the proximal entropy formulation itself, which measures each token's entropy relative to its neighbors; reflecting the sequence at the boundary effectively treats nearby tokens on the opposite side of $t$ as the missing neighbors, consistent with the local-context interpretation of $e_{i,t}$.

Boundary effects are confined to roughly $W/T_i \approx 5\%$ of tokens for our default $W=101$ and typical sequence lengths, and we found this choice to work well empirically.

\subsection{Software and Compute}
\label{app:compute}

All experiments are run on 2 NVIDIA H200 GPUs. We use the ROLL framework for the main experiments and the VeRL framework for single-stream experiments, with vLLM for rollout generation.

\section{Runtime Analysis}
\label{app:runtime}

To verify that proximal entropy weighting adds negligible overhead, we measure the total wall-clock training time of PEPO and the baselines under identical hardware (2$\times$H200), framework (ROLL), and training configuration (500 steps, $G=8$, max length 4096). Table~\ref{tab:runtime} reports the results for the Qwen3-1.7B and Qwen3-4B.

\begin{table}[H]
\centering
\caption{Total wall-clock training time on 2$\times$H200 GPUs. All methods are run with identical training configurations for 500 steps. PEPO adds negligible overhead over GRPO across both models.}
\label{tab:runtime}
\begin{tabular}{lcc}
\toprule
\textbf{Method} & \textbf{Qwen3-1.7B} & \textbf{Qwen3-4B} \\
\midrule
GRPO            & 1d 7h 32m    & 1d 19h 25m \\
80/20           & 1d 7h 28m    & 1d 20h 14m \\
Entropy Adv.    & 1d 8h 28m    & 1d 21h 3m \\
PEPO (Ours)     & 1d 7h 54m & 1d 19h 45m \\
\bottomrule
\end{tabular}
\end{table}

As shown in Table~\ref{tab:runtime}, all entropy-based methods (80/20, Entropy Adv., PEPO) run within minutes of GRPO across both models, well within the margin of variation expected from differences in random seeds and hardware scheduling.

\section{Additional Bias Analyses of Proximal Entropy}
\label{app:bias}

In Section~2.2 of the main paper, we showed that proximal entropy is empirically invariant to prompt difficulty (Table~1) and positional trends (Figure~2). In this section, we provide additional analyses showing that this invariance is robust across window sizes and is not an artifact of the softmax operation.

\begin{figure}[H]
\centering
\includegraphics[width=0.8\linewidth]{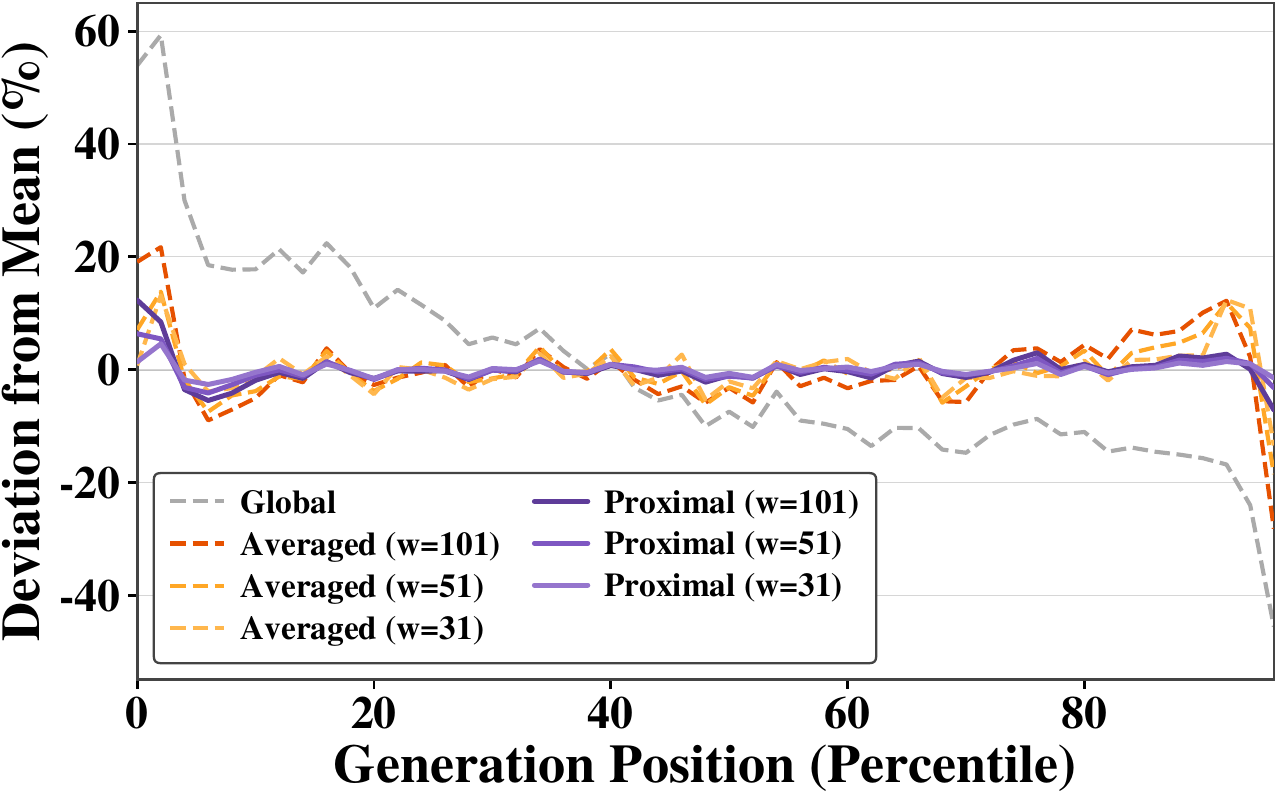}
\caption{Deviation of windowed entropy statistics from their sequence-level mean across generation percentiles, for window sizes $W \in \{31, 51, 101\}$. We compare two aggregation choices: proximal entropy (solid purple), defined as a softmax over the window, and a simple windowed average (dashed orange). Both yield flat curves across all three window sizes, in contrast to global entropy (gray dashed) which decays substantially. This shows that the positional invariance is robust to the window size and is a property of the windowing itself rather than of the softmax operation. Computed over 992 rollouts on the validation set with Llama-3.2-3B-Instruct.}
\label{fig:window_bias}
\end{figure}

\paragraph{Invariance across window sizes and aggregation choices.}
Proposition~1(ii) establishes that proximal entropy is approximately invariant to positional trends for sufficiently small $W$. To empirically verify this, we plot the deviation of two window-based statistics from the sequence-level mean across generation percentiles for $W \in \{31, 51, 101\}$: (i) proximal entropy as defined in Eq.~6 and (ii) a simple windowed average of token entropies without the softmax operation. As shown in Figure~\ref{fig:window_bias}, both statistics yield flat curves across all three window sizes, in contrast to global entropy (gray dashed) which decays substantially over the course of generation.

This result has two implications. First, the positional invariance of proximal entropy is robust to the choice of $W$, holding across the range of window sizes considered in our experiments. Second, the mitigation of positional bias is a property of the local windowing itself rather than the softmax operation: even a simple windowed average suffices to remove the positional trend. The softmax in PEPO serves a different purpose, which is to convert a window of entropies into a continuous, normalized per-token weight that emphasizes tokens with relatively higher entropy than their local neighbors. The two design choices (windowing and softmax-based weighting) are therefore separable, and the bias-removal property attaches to the former.

\section{Additional Evaluation and Ablations}
\label{app:additional_experiments}

\subsection{Held-Out Code Generation: LiveCodeBench}
\label{app:livecodebench}

To test transfer beyond mathematical reasoning, we evaluate the Qwen3-4B GRPO, 80/20, and PEPO checkpoints on the held-out \texttt{release\_v6/code\_generation\_lite} split of LiveCodeBench, which contains 1,055 problems. LiveCodeBench was not included in RL training; the ROLL training mixture did include separate verifiable coding problems from KodCode. The three checkpoints share the same base model, training data, and training budget. We use the official LiveCodeBench evaluation implementation (\url{https://github.com/LiveCodeBench/LiveCodeBench}) with temperature 0.2, top-$p$ 0.95, a maximum of 4,096 generated tokens, thinking mode enabled, and bfloat16 inference. Table~\ref{tab:livecodebench_summary} reports overall pass@1, and Table~\ref{tab:livecodebench_breakdown} reports the difficulty and platform breakdowns.

PEPO reaches 46.35\% pass@1, a 14.60 percentage-point improvement over GRPO and 5.31 points over 80/20. Its gains over GRPO are positive across all three difficulty groups and all listed platforms. The Codeforces subset contains only nine problems, so that row should be interpreted cautiously.

\begin{table}[H]
\centering
\small
\caption{Held-out LiveCodeBench \texttt{release\_v6/code\_generation\_lite} pass@1 point estimates for Qwen3-4B.}
\label{tab:livecodebench_summary}
\begin{tabular}{lrrr}
\toprule
Method & Correct & Pass@1 & Gain vs. GRPO \\
\midrule
GRPO & 335 / 1,055 & 31.75\% & --- \\
80/20 & 433 / 1,055 & 41.04\% & +9.29 pp \\
PEPO & 489 / 1,055 & \textbf{46.35\%} & \textbf{+14.60 pp} \\
\bottomrule
\end{tabular}
\end{table}

\begin{table}[H]
\centering
\small
\caption{LiveCodeBench pass@1 breakdown by difficulty and problem platform. The Codeforces subset contains only 9 problems.}
\label{tab:livecodebench_breakdown}
\setlength{\tabcolsep}{8pt}
\begin{tabular}{lrrrrr}
\toprule
Subset & Problems & GRPO & 80/20 & PEPO & PEPO $-$ GRPO \\
\midrule
\multicolumn{6}{l}{\textit{By difficulty}} \\
Easy & 322 & 76.09\% & 84.78\% & \textbf{90.99\%} & +14.91 pp \\
Medium & 383 & 22.72\% & 37.60\% & \textbf{43.86\%} & +21.15 pp \\
Hard & 350 & 0.86\% & 4.57\% & \textbf{8.00\%} & +7.14 pp \\
\addlinespace
\multicolumn{6}{l}{\textit{By platform}} \\
AtCoder & 602 & 31.23\% & 38.87\% & \textbf{42.86\%} & +11.63 pp \\
LeetCode & 444 & 32.66\% & 43.92\% & \textbf{50.68\%} & +18.02 pp \\
Codeforces & 9 & 22.22\% & 44.44\% & \textbf{66.67\%} & +44.44 pp \\
\bottomrule
\end{tabular}
\end{table}

\subsection{Weighting Design Ablation}
\label{app:weighting_ablation}

We decompose the Qwen3-4B mean accuracy gain by changing one component at a time: global to proximal entropy for binary selection, binary selection to continuous weighting over all trajectories, and then positive-only application. Table~\ref{tab:weighting_ablation} reports the aggregate results.

\begin{table}[H]
\centering
\small
\caption{Sequential weighting ablation on Qwen3-4B. Mean accuracy averages AIME 2024, AIME 2025, AMC, and MATH500; the first gain averages the benchmark-level differences in Table~3.}
\label{tab:weighting_ablation}
\setlength{\tabcolsep}{4pt}
\begin{tabular}{@{}p{0.62\linewidth}rr@{}}
\toprule
Configuration & Mean & Gain (pp) \\
\midrule
Global entropy, binary selection (80/20) & 55.74 & --- \\
Proximal entropy, binary selection & 56.87 & +1.14 \\
Proximal entropy, continuous weighting on all rollouts & 57.53 & +0.66 \\
Proximal entropy, continuous weighting on positive rollouts & \textbf{58.01} & +0.48 \\
\bottomrule
\end{tabular}
\end{table}

Localization adds 1.14 points, moving from binary to continuous weighting over all trajectories adds 0.66 points, and restricting continuous weighting to positive-reward rollouts adds a further 0.48 points. The first gain is the average of the four benchmark-level changes shown in Table~3; the all-rollout continuous-weighting result was reported without run-level uncertainty, so its comparison with positive-only weighting is descriptive. Even with all-rollout weighting, the mean remains 7.38 points above GRPO (57.53 versus 50.15).

\subsection{Temperature Sensitivity}
\label{app:temperature_sensitivity}

We vary the proximal-entropy softmax temperature $\tau$ on Llama-3.2-3B-Instruct while keeping the remaining settings fixed. Table~\ref{tab:temperature_sensitivity} gives the reported point estimates.

\begin{table}[H]
\centering
\small
\caption{Temperature sensitivity for Llama-3.2-3B-Instruct. Results are the reported point estimates; no seed-level uncertainty was provided for this sweep. Best result per column is bolded.}
\label{tab:temperature_sensitivity}
\setlength{\tabcolsep}{5pt}
\begin{tabular}{crrrrr}
\toprule
$\tau$ & AIME24 & AIME25 & AMC & MATH500 & Mean \\
\midrule
0.5 & 8.75 & 1.25 & 23.72 & \textbf{49.40} & 20.78 \\
1.0 & \textbf{9.10} & \textbf{1.46} & \textbf{24.12} & 49.33 & \textbf{21.00} \\
2.0 & 8.54 & \textbf{1.46} & 23.72 & 49.00 & 20.68 \\
\bottomrule
\end{tabular}
\end{table}

Mean accuracy spans only 20.68--21.00 across the sweep, with the default $\tau=1$ giving the highest reported mean.

\clearpage
\section*{Per-Run Experimental Results}
\subsection*{Main Results (Table 2)}

\begin{table}[H]
\centering
\caption{Per-run accuracy for the main results in Table~2. Each row reports the avg@1 accuracy on MATH500 and avg@16 accuracy on AMC, AIME 2024, and AIME 2025 for a single run. Mean rows match the values reported in Table~2.}
\label{tab:per_run_main}
\small
\begin{tabular}{llccccc}
\toprule
\textbf{Model} & \textbf{Method} & \textbf{Run} & \textbf{AIME2024} & \textbf{AIME2025} & \textbf{AMC} & \textbf{MATH500} \\
\midrule
\multirow{16}{*}{Qwen3-1.7B}
& \multirow{4}{*}{GRPO}            & 1    & 16.42 & 18.26 & 51.22 & 79.20 \\
&                                  & 2    & 17.07 & 19.36 & 50.38 & 81.20 \\
&                                  & 3    & 16.24 & 18.33 & 49.25 & 80.40 \\
&                                  & Mean & 16.58 & 18.65 & 50.28 & 80.27 \\
\cmidrule(lr){2-7}
& \multirow{4}{*}{Entropy Adv.}    & 1    & 22.25 & 20.53 & 51.61 & 79.80 \\
&                                  & 2    & 19.37 & 17.50 & 53.69 & 82.60 \\
&                                  & 3    & 22.71 & 20.41 & 52.26 & 81.80 \\
&                                  & Mean & 21.44 & 19.48 & 52.52 & 81.40 \\
\cmidrule(lr){2-7}
& \multirow{4}{*}{80/20}           & 1    & 22.29 & 21.46 & 51.95 & 79.20 \\
&                                  & 2    & 21.25 & 21.04 & 52.63 & 80.80 \\
&                                  & 3    & 19.37 & 20.00 & 51.80 & 82.20 \\
&                                  & Mean & 20.97 & 20.83 & 52.13 & 80.73 \\
\cmidrule(lr){2-7}
& \multirow{4}{*}{PEPO}            & 1    & 20.84 & 22.71 & 54.54 & 82.40 \\
&                                  & 2    & 22.91 & 21.67 & 56.54 & 83.00 \\
&                                  & 3    & 21.46 & 22.91 & 54.69 & 82.60 \\
&                                  & Mean & 21.74 & 22.43 & 55.26 & 82.67 \\
\midrule
\multirow{16}{*}{Qwen3-4B}
& \multirow{4}{*}{GRPO}            & 1    & 32.68 & 22.92 & 60.25 & 83.00 \\
&                                  & 2    & 33.33 & 24.38 & 59.18 & 85.40 \\
&                                  & 3    & 32.25 & 23.31 & 60.45 & 84.60 \\
&                                  & Mean & 32.75 & 23.54 & 59.96 & 84.33 \\
\cmidrule(lr){2-7}
& \multirow{4}{*}{Entropy Adv.}    & 1    & 36.41 & 24.88 & 68.25 & 89.60 \\
&                                  & 2    & 37.08 & 26.46 & 67.18 & 90.00 \\
&                                  & 3    & 34.15 & 27.29 & 68.06 & 89.60 \\
&                                  & Mean & 35.88 & 26.21 & 67.83 & 89.73 \\
\cmidrule(lr){2-7}
& \multirow{4}{*}{80/20}           & 1    & 32.50 & 30.83 & 67.77 & 88.40 \\
&                                  & 2    & 36.87 & 30.62 & 68.46 & 90.40 \\
&                                  & 3    & 36.67 & 30.42 & 65.53 & 90.40 \\
&                                  & Mean & 35.35 & 30.62 & 67.25 & 89.73 \\
\cmidrule(lr){2-7}
& \multirow{4}{*}{PEPO}            & 1    & 38.13 & 31.88 & 70.17 & 91.80 \\
&                                  & 2    & 40.63 & 31.88 & 69.58 & 89.40 \\
&                                  & 3    & 38.96 & 31.66 & 68.99 & 93.00 \\
&                                  & Mean & 39.24 & 31.81 & 69.58 & 91.40 \\
\midrule
\multirow{16}{*}{Llama-3.2-3B-Instruct}
& \multirow{4}{*}{GRPO}            & 1    & 6.46  & 0.83  & 21.69 & 48.00 \\
&                                  & 2    & 7.08  & 1.46  & 20.78 & 48.80 \\
&                                  & 3    & 8.12  & 0.42  & 21.83 & 46.20 \\
&                                  & Mean & 7.22  & 0.90  & 21.43 & 47.67 \\
\cmidrule(lr){2-7}
& \multirow{4}{*}{Entropy Adv.}    & 1    & 8.12  & 0.42  & 24.24 & 47.60 \\
&                                  & 2    & 9.17  & 0.42  & 23.86 & 48.20 \\
&                                  & 3    & 7.50  & 0.83  & 22.81 & 47.00 \\
&                                  & Mean & 8.26  & 0.56  & 23.64 & 47.60 \\
\cmidrule(lr){2-7}
& \multirow{4}{*}{80/20}           & 1    & 7.92  & 0.42  & 23.26 & 47.60 \\
&                                  & 2    & 9.12  & 0.62  & 21.53 & 48.20 \\
&                                  & 3    & 8.75  & 0.42  & 22.91 & 47.20 \\
&                                  & Mean & 8.60  & 0.49  & 22.57 & 47.67 \\
\cmidrule(lr){2-7}
& \multirow{4}{*}{PEPO}            & 1    & 8.54  & 1.67  & 23.71 & 49.40 \\
&                                  & 2    & 9.58  & 0.83  & 24.40 & 49.40 \\
&                                  & 3    & 9.17  & 1.87  & 24.24 & 49.20 \\
&                                  & Mean & 9.10  & 1.46  & 24.12 & 49.33 \\
\bottomrule
\end{tabular}
\end{table}

\subsection*{Proximal Entropy as a General Formulation (Table 3)}

\begin{table}[H]
\centering
\caption{Per-run accuracy for the drop-in proximal entropy results in Table~3. Mean rows match the values reported in Table~3.}
\label{tab:per_run_dropin}
\small
\begin{tabular}{llccccc}
\toprule
\textbf{Model} & \textbf{Method} & \textbf{Run} & \textbf{AIME2024} & \textbf{AIME2025} & \textbf{AMC} & \textbf{MATH500} \\
\midrule
\multirow{4}{*}{Qwen3-1.7B}
& \multirow{4}{*}{80/20 + Prox.}   & 1    & 22.71 & 20.83 & 53.16 & 82.60 \\
&                                  & 2    & 21.04 & 20.42 & 52.71 & 82.20 \\
&                                  & 3    & 22.29 & 20.83 & 54.22 & 82.00 \\
&                                  & Mean & 22.01 & 20.69 & 53.36 & 82.27 \\
\midrule
\multirow{4}{*}{Qwen3-4B}
& \multirow{4}{*}{80/20 + Prox.}   & 1    & 38.75 & 31.04 & 69.65 & 90.60 \\
&                                  & 2    & 33.54 & 31.88 & 70.63 & 90.40 \\
&                                  & 3    & 36.25 & 31.88 & 67.85 & 90.00 \\
&                                  & Mean & 36.18 & 31.60 & 69.38 & 90.33 \\
\bottomrule
\end{tabular}
\end{table}

\subsection*{Single-Stream RL (Table 4)}

\begin{table}[H]
\centering
\caption{Per-run accuracy for the single-stream RL results in Table~4 on Qwen3-4B. Mean rows match the values reported in Table~4.}
\label{tab:per_run_pespo}
\small
\begin{tabular}{lccccc}
\toprule
\textbf{Method} & \textbf{Run} & \textbf{AIME2024} & \textbf{AIME2025} & \textbf{AMC} & \textbf{MATH500} \\
\midrule
\multirow{4}{*}{SPO}               & 1    & 37.92 & 31.25 & 70.71 & 90.00 \\
                                   & 2    & 35.00 & 31.04 & 72.14 & 90.80 \\
                                   & 3    & 34.17 & 31.88 & 69.58 & 91.00 \\
                                   & Mean & 35.70 & 31.39 & 70.81 & 90.60 \\
\midrule
\multirow{4}{*}{S-80/20}           & 1    & 36.46 & 26.67 & 68.98 & 90.60 \\
                                   & 2    & 35.21 & 28.75 & 70.56 & 91.40 \\
                                   & 3    & 34.17 & 27.50 & 70.63 & 91.00 \\
                                   & Mean & 35.28 & 27.64 & 70.06 & 91.00 \\
\midrule
\multirow{4}{*}{S-Entropy Adv.}    & 1    & 38.75 & 22.29 & 68.30 & 89.80 \\
                                   & 2    & 40.21 & 22.08 & 68.90 & 89.40 \\
                                   & 3    & 37.71 & 21.04 & 68.07 & 89.20 \\
                                   & Mean & 38.89 & 21.80 & 68.42 & 89.47 \\
\midrule
\multirow{4}{*}{PESPO}             & 1    & 41.25 & 31.04 & 71.76 & 90.60 \\
                                   & 2    & 42.29 & 32.50 & 72.74 & 90.40 \\
                                   & 3    & 41.67 & 31.25 & 70.63 & 91.00 \\
                                   & Mean & 41.74 & 31.60 & 71.71 & 90.67 \\
\bottomrule
\end{tabular}
\end{table}

\subsection*{Window Size Ablation (Table 6)}

\begin{table}[H]
\centering
\caption{Per-run accuracy for the window size ablation in Table~6 on Qwen3-4B. Mean rows match the values reported in Table~6.}
\label{tab:per_run_window}
\small
\begin{tabular}{lccccc}
\toprule
\textbf{Window Size} & \textbf{Run} & \textbf{AIME2024} & \textbf{AIME2025} & \textbf{AMC} & \textbf{MATH500} \\
\midrule
\multirow{4}{*}{$W = 51$}    & 1    & 34.59 & 30.20 & 68.36 & 90.20 \\
                             & 2    & 37.50 & 29.59 & 67.82 & 91.00 \\
                             & 3    & 36.67 & 32.08 & 68.99 & 90.80 \\
                             & Mean & 36.25 & 30.62 & 68.39 & 90.67 \\
\midrule
\multirow{4}{*}{$W = 151$}   & 1    & 37.08 & 27.08 & 69.87 & 92.00 \\
                             & 2    & 36.67 & 29.59 & 68.89 & 92.00 \\
                             & 3    & 36.67 & 28.13 & 69.29 & 92.80 \\
                             & Mean & 36.81 & 28.27 & 69.35 & 92.27 \\
\bottomrule
\end{tabular}
\end{table}

\end{document}